%% file: paper.tex
\documentclass[runningheads]{llncs}

\input{paper_packages}
\input{paper_commands}

\begin{document}
\pagestyle{headings}
\mainmatter

\title{Novel-View Human Action Synthesis} 
\titlerunning{Novel-View Human Action Synthesis}

\author{Mohamed Ilyes Lakhal\inst{1} \and
Davide Boscaini\inst{2} \and
Fabio Poiesi\inst{2} \and
Oswald Lanz\inst{2} \and
Andrea Cavallaro\inst{1}}

\authorrunning{M.I. Lakhal et al.}

\institute{Centre for Intelligent Sensing, Queen Mary University of London, UK \and
Technologies of Vision, Fondazione Bruno Kessler, Italy
\email{\{m.i.lakhal,a.cavallaro\}@qmul.ac.uk}, \email{\{dboscaini,poiesi,lanz\}@fbk.eu\\}}

\maketitle
\input{sec_abstract}
\input{sec_intro}
\input{sec_background}
\input{sec_method}
\input{sec_exp}
\input{sec_conclusion}
\input{sec_supp}
\clearpage
\bibliographystyle{splncs}
\bibliography{refs}
\end{document}

%% file: paper_packages.tex
\usepackage{amsmath,amssymb}
\usepackage{tikz}
\usepackage{xcolor,colortbl}
\usepackage{microtype}
\usepackage{booktabs}
\usepackage{multirow}
\usepackage{caption}
\usepackage{subcaption}
\usepackage{algorithm,algpseudocode}
\algnewcommand{\LineComment}[1]{\State \(\triangleright\) #1}
\usepackage[percent]{overpic}
\usepackage{pict2e}
\usepackage{enumitem}

\usepackage{hyperref}
\hypersetup{
    colorlinks=true,
    linkcolor=blue,
    filecolor=magenta,      
    urlcolor=cyan,
}

%% file: paper_commands.tex
\newcommand{\etal}{\emph{et al. }}
\newcommand{\norm}[1]{\left\lVert#1\right\rVert}
\definecolor{Gray}{gray}{0.9}

%% file: sec_abstract.tex
\begin{abstract}
Novel-View Human Action Synthesis aims to synthesize the movement of a body from a virtual viewpoint, given a video from a real viewpoint.
We present a novel 3D reasoning to synthesize the target viewpoint. 
We first estimate the 3D mesh of the target body and transfer the rough textures from the 2D images to the mesh.
As this transfer may generate sparse textures on the mesh due to frame resolution or occlusions.
We produce a semi-dense textured mesh by propagating the transferred textures both locally, within local geodesic neighborhoods, and globally, across symmetric semantic parts.
Next, we introduce a context-based generator to learn how to correct and complete the residual appearance information.
This allows the network to independently focus on learning the  foreground and background synthesis tasks.
We validate the proposed solution on the public NTU RGB+D dataset. The code and resources are available at \url{https://bit.ly/36u3h4K}.
\end{abstract}

%% file: sec_intro.tex
\section{Introduction}
Novel-view human action synthesis is the problem of reproducing a person performing an action from a virtual viewpoint~\cite{Lakhal_2019_ICCV}.
The ability to synthesize of one or more novel viewpoints of an action is attractive for extended reality~\cite{Bertel_2019_TVCG}, action recognition~\cite{Li_2018_NIPS} and free-viewpoint video~\cite{Rematas_2018_CVPR}.

Recent works~\cite{Natsume_2019_CVPR,Saito_2019_ICCV,Lombardi_2019_TOG,Thies_2020_ICLR} have shown the ability to synthesize high-quality images, but with limiting assumptions on the input data. 
SiCloPe~\cite{Natsume_2019_CVPR} takes a frontal input image and uses canonical views to reconstruct the 3D mesh through supervision. However, using a ground-truth mesh from real images is not a realistic assumption.
Similarly, PIFu~\cite{Saito_2019_ICCV} predicts a dense 3D occupancy field using multiple input views. 
This method expects high-resolution ground-truth mesh and a neutral background which hinder generalization to real-world scenarios when  backgrounds are cluttered. 
The method proposed in~\cite{Weng_2019_CVPR} creates an animated version of an image containing a person in the center. 
An initial mesh is first estimated and then corrected. 
However, the mesh construction part is computationally costly and incompatible with the extension of the model to videos. 
Furthermore, the texture filling is based on heuristics or requires human intervention.
If multiple real views of the same scene are available, the rendering of an arbitrary virtual view can be successfully addressed~\cite{Mustafa_2017_CVPR,Bansal_2020_CVPR}. For example, the method proposed in Bansal \etal~\cite{Bansal_2020_CVPR} combines the information available from multiple camera views to reconstruct the geometry of a static scene. Then, a neural network based model is used to compose the dynamics on top of the static scene.
However, with only a single video (view) as input the problem becomes much more challenging and largely unexplored. 
To the best of our knowledge, VDNet~\cite{Lakhal_2019_ICCV} is the only previous work addressing it.

Thanks to the rapid development on human mesh recovery~\cite{Bogo_ECCV_2016,Kanazawa_2018_CVPR,Kanazawa_2019_CVPR,Kolotouros_2019_ICCV,Pavlakos_2019_ICCV}, we can obtain 3D representations from images or videos. Our approach consists of a two-stage pipeline.
In the first stage we exploit a novel 3D reasoning to produce a sparse initialization for the virtual view.
In the second stage we introduce Geometric texture Transfer Network (GTNet), a context-based generator that aims to correct and complete such initial guess by learning the residual appearance information.
For each frame captured from the real view we estimate the 3D mesh of the human actor using \cite{Kanazawa_2019_CVPR} where the parameters of the Skinned Multi-Person Linear (SMPL) model \cite{Loper_2015_TOG} are learned to morph a canonical 3D model of the human body to fit the 2D projection of the human actor pose and shape.
Given such 3D model, we transfer the appearance information from the 2D video to the 3D mesh. 
This results in a sparse texture on the 3D mesh because of occlusions. 
We propose to compute the missing information by exploiting the knowledge of the 3D model both at a local and global scale. 
Locally, missing values within a geodesic neighborhood are computed by interpolating the input sparse texture. 
More globally, if a part of the 3D model (e.g. an arm or a leg) lacks texture information but its symmetric counterpart contains it, we propagate it.
The texture on the mesh (in 3D) obtained in this way is then projected (rendered) on the novel view (in 2D). 
The estimated 3D model thus acts as a proxy to transfer appearance information from the input (real) view to the target (virtual) view.
The design of our approach is inspired by pixel warping methods~\cite{Leonid_2011_CVPR,Feng_2011_TOG} that create realistic human images from existing frames (or views). Differently from VDNet~\cite{Lakhal_2019_ICCV}, we exploit the geometric properties of the input prior to facilitate the transfer to the target view.
Unlike motion-transfer methods (\textit{e.g.}~\cite{Siarohin_2019_NIPS}), we learn the geometry and the appearance of a novel (virtual) view.

%% file: sec_background.tex
\section{Related work} \label{sec:related}
Methods for novel-view synthesis that focus on humans can be based on computer graphics, learning, or combining 3D mesh representations and learning. 
These methods are discussed in this section jointly with a discussion on the importance of the modality used to synthesize the novel-view.

\textbf{Novel-View Image Synthesis.} Graphics based methods \cite{Chen_2019_ICCV,Saito_2019_ICCV,Bhatnagar_2019_ICCV} rely on the abundance of ground-truth data to achieve high quality synthesis. For example \cite{Alldieck_2019_CVPR,Bhatnagar_2019_ICCV,Alldieck_2018_CVPR} use image or sequence of frames to learn the displacement of clothing on top of the SMPL~\cite{Loper_2015_TOG} model. Differently, the methods in~\cite{Saito_2019_ICCV,Natsume_2019_CVPR} use high quality human mesh representations from the Renderpeople dataset\footnote{\url{https://renderpeople.com/}, accessed September 2020}. These representations enable the model to achieve high quality results, but fail to generalize in uncontrolled setups and need a few viewpoints, which may be hard to obtain, to perform the synthesis. 

Learning (or data-driven) approaches \cite{Lakhal_2019_ICCV,Li_2018_NIPS,Ma_2017_NIPS,zhao_2018_ACMMM} use spatial cues about the human subject to synthesize the target view. A drawback of such approaches is the poor generalization to unseen views and the difficulty to handle occlusions. 

A new direction of work considers the use of a 3D model estimated directly from raw images~\cite{Zanfir_2019_AAAI,Liu_2019_ICCV,Li_2019_CVPR}. Liu \textit{et al.}~\cite{Liu_2019_ICCV} enforce feature warping of the input view in the network structure to synthesize the novel view.

\textbf{Video Synthesis.} We categorise the methods solving this problem into two classes: unconstrained or constrained synthesis. The first category tries to learn the distribution of the data during training. The video is therefore a sample from the learned distribution~\cite{Saito_2017_ICCV,Tulyakov_2018_CVPR,Vondrick_2016_NIPS}. Since the datasets available are most often a sparse representation of the distribution of the true data, the generated videos generally are limited to few applications. The constrained video synthesis~\cite{Siarohin_2019_NIPS,Ceyuan_2018_ECCV,wang_2019_NIPS} relies on context (\textit{e.g.} image sequence~\cite{Siarohin_2019_NIPS}) or spatial cues (\textit{e.g.} keypoints~\cite{Ceyuan_2018_ECCV,wang_2019_NIPS}). Applications include action imitation~\cite{Siarohin_2019_NIPS} and video prediction~\cite{Li_2018_ECCV}. 

\textbf{Novel-View Video Synthesis.} Recently, Lakhal \textit{et al.}~\cite{Lakhal_2019_ICCV} introduced the task of {\em novel-view video synthesis} which shares the challenges of both the novel view synthesis (\textit{i.e} dealing with occlusion) and video synthesis (\textit{i.e} maintaining temporal consistency across frames). The assumptions are the availability of only one input view and the modalities about the target view which can be either given or computed. Furthermore, the problem is different from the pose-guided human image synthesis~\cite{Ma_2017_NIPS} where the background synthesis is not taken into account, the pose is not constrained to the view (\textit{i.e.} cannot model the 3D structure of the scene), and these methods fail to maintain temporal consistency.  In this paper, we show that by estimating the texture we can approximate the target feature with a simple mapping. Using the proposed context-based architecture, the network can focus on the background synthesis. We exploit the 3D mesh information and use the input-target view association as guidance in the novel-view synthesis process. Also, we explicitly handle the temporal consistency of the synthesized frames. Furthermore, we handle  self occlusions using visible information and transfer it to neighboring occluded parts.

\textbf{Prior modalities.} Deep learning based methods made progress on estimating accurate modalities (\textit{e.g.} depth and 2D/3D keypoints) from object priors (\textit{e.g.} human). This includes human pose estimation~\cite{Raaj_2019_CVPR}, human part segmentation~\cite{Yang_2019_CVPR}, or human mesh recovery~\cite{Kanazawa_2019_CVPR,Bogo_ECCV_2016}. 
The performance of a neural network-based generator for novel-view synthesis relies heavily on the modalities derived from the prior used about the target view (we only consider human priors). Early works rely on $2$D keypoints of the human body joints~\cite{Siarohin_2018_CVPR,Ma_2017_NIPS,Pumarola_2018_CVPR,Liqian_2018_CVPR,Chan_2019_ICCV,Esser_2018_CVPR,Balakrishnan_2018_CVPR,Qian_2018_ECCV}.

The skeleton indicates the spatial location of the person on the target view. The network then has to learn how to extract and transfer appearance information from the input view to generate the target image. A segmentation map could be considered as another modality~\cite{Raj_2018_ECCV,Dong_2018_NIPS}.
DensePose~\cite{Guler_2018_CVPR} maps the pixels of an RGB image of human to a 3D surface and is used in~\cite{Dong_2018_NIPS,Neverova_2018_ECCV}.
Li \etal \cite{Li_2019_CVPR} proposed to represent the pose as a rendered 3D body mesh. 

%% file: sec_method.tex
\begin{figure}[t!]
    \centering
    \includegraphics[width=1.\linewidth]{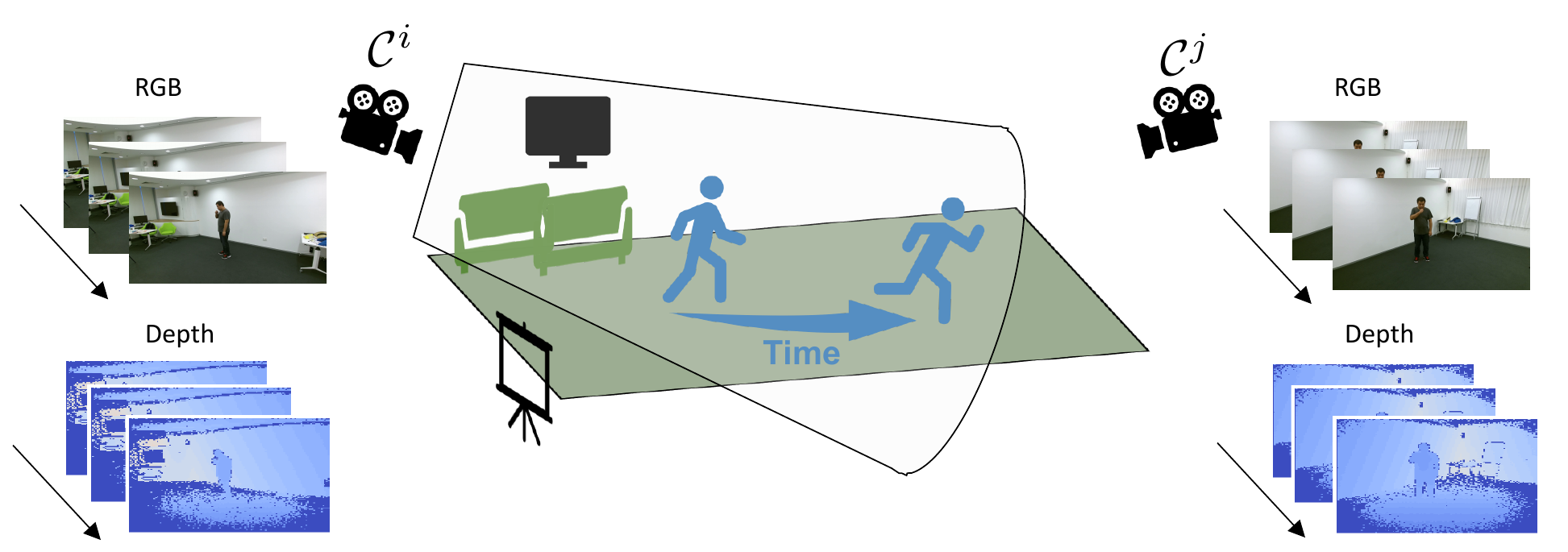}
    \caption{Given a video of a person performing an action recorded from an input view $\mathcal{C}^i$, we synthesize how it would appear from a target (or virtual) view $\mathcal{C}^j$. Each view can be captured with a set of modalities (e.g. RGB, depth, skeleton).}
    \label{fig:pf}
\end{figure} 

\section{Method}
\subsection{Problem definition} 
\label{sec:problem}

Let $\mathcal{C} = \{ \mathcal{C}^i \}_{i = 1}^{V}$ be $V$ static cameras (views) placed at different positions in a scene. 
Each view $i \in \{1, \dots, V \}$ is represented as a sequence of RGB images $x^i_t \in \mathbb{R}^{w \times h \times 3}$ of width $w$ and height $h$ pixels and indexed by the timestep $t = 1, \dots, T$.
The sequence $x^i = \{x^i_t\}_{t = 1}^T$ is an instance of the scene captured from the input view camera $\mathcal{C}^i$.
Each view contains $M$ different modalities $p^i = \{p^i_1, \dots, p^i_M \}$ \textit{e.g.} depth and skeleton (see Fig.~\ref{fig:pf}).
Each modality has to at least spatially localise the person in the scene (\textit{i.e.,} foreground).
The modalities of the virtual view are computed by transforming $p^i$ with the information in both $\mathcal{C}^i$ and $\mathcal{C}^j$.

The aim of novel-view video synthesis is to reconstruct the RGB sequence $x^j$ of a scene from an input view to a target view. Given a parametric function $G_{i \to j}$ called \emph{generator} we have: $x^j = G_{i \to j}(x^i, p^j)$.

\subsection{3D human body prior}
\begin{figure}[t!]
    \centering
    \begin{overpic}[width=\linewidth]{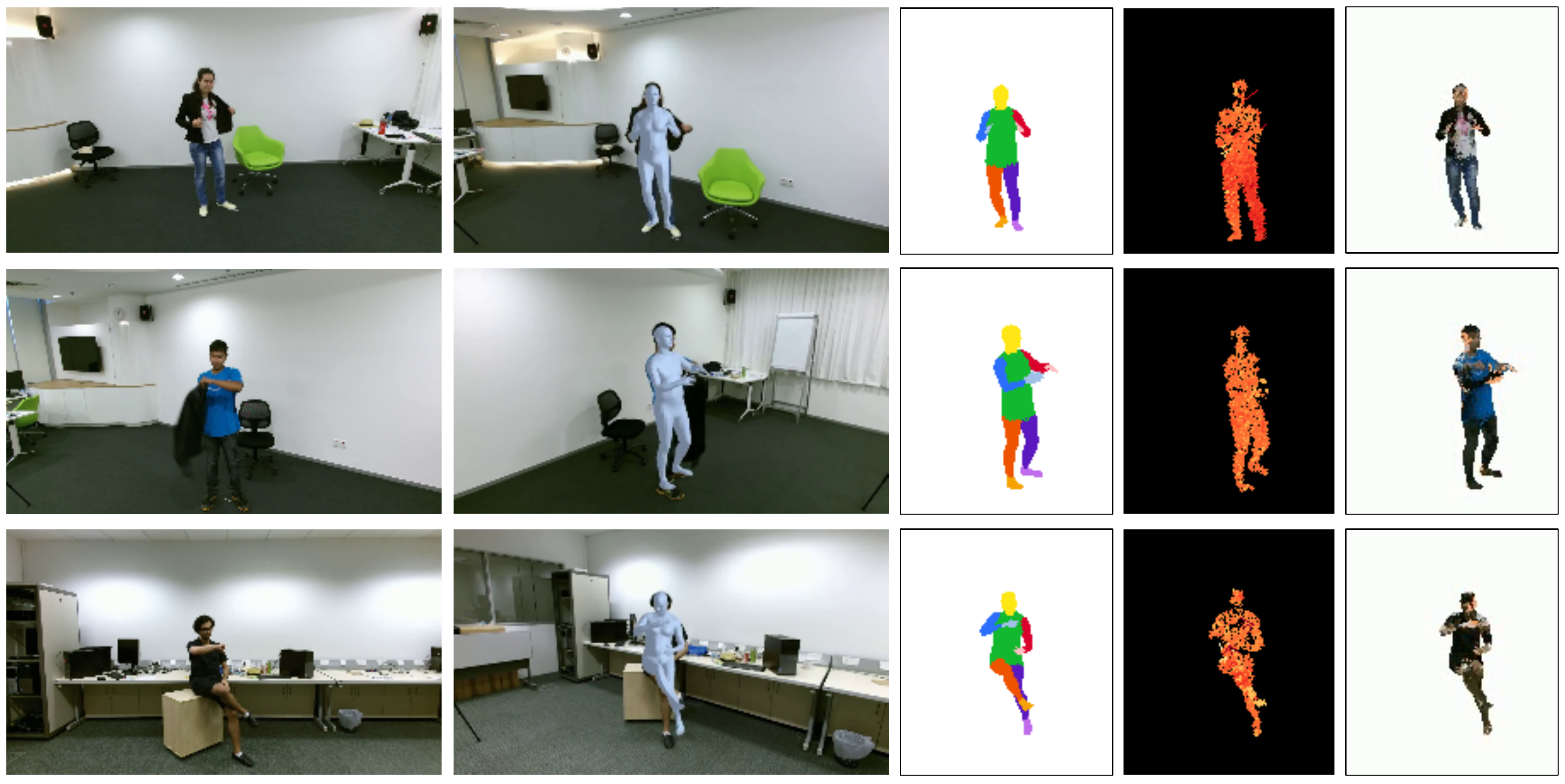}
        \put(13,-2){\color{black}\scriptsize (a)}
        \put(41,-2){\color{black}\scriptsize (b)}
        \put(63,-2){\color{black}\scriptsize (c)}
        \put(77,-2){\color{black}\scriptsize (d)}
        \put(91,-2){\color{black}\scriptsize (e)}

        \put(13,51){\color{black}\scriptsize $x^i$}
        \put(39,51){\color{black}\scriptsize $x^j + \mathcal{M}^j$}
        \put(63.4,51){\color{black}\scriptsize $\mathcal{S}^j$}
        \put(74.5,51){\color{black}\scriptsize $\mathcal{O}^j_{t+1 \to t}$}
        \put(91,51){\color{black}\scriptsize $\mathcal{T}_{i \to j}$}
    \end{overpic}
    \caption{Target view modalities obtained with the 3D human body prior extracted from NTU-RGB+D dataset~\cite{Shahroudy_2016_CVPR}: (a) input view; (b) target view with the rendered mesh; (c) segmentation map; (d) foreground motion; (e) texture transfer.}
    \label{fig:mods_}
\end{figure}
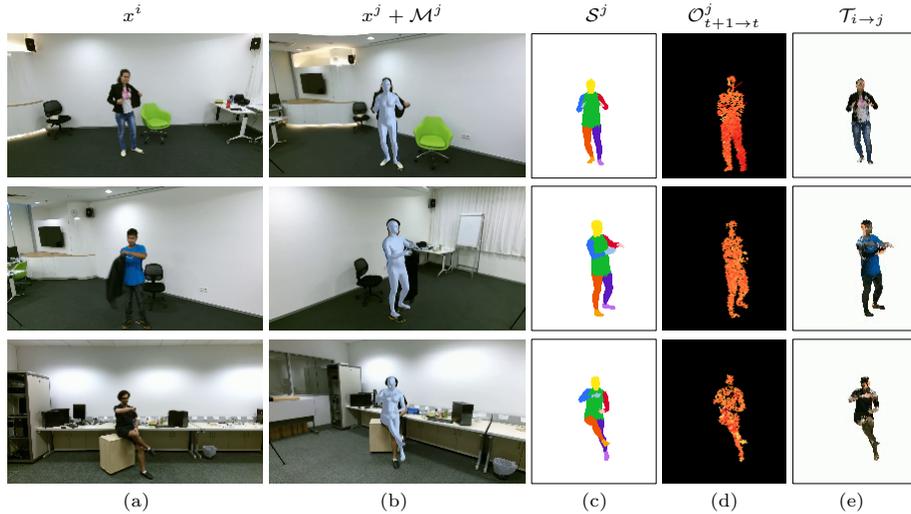

We model the foreground using a 3D mesh assuming that the video stream contains only humans. 
We use the estimated 3D mesh as a proxy to transfer the appearance information from the input view to the novel view.
In this section, we show that we can compute other modalities of the target view by fully exploiting the one-to-one correspondence between vertices of the 3D mesh (Fig.~\ref{fig:mods_}).

\noindent\textbf{Foreground prior.}
We model the foreground (\textit{i.e.} the target subject) using a geometric approach that exploits one-to-one correspondences between the vertices of the 3D meshes from the real and virtual views.
As in \cite{Kanazawa_2019_CVPR}, we model the human body using the SMPL model \cite{Loper_2015_TOG}. A set of three connected vertices defines a face on the 3D mesh. The SMPL model is composed of $N_f = 13776$ faces that are uniquely identified by a face map $\mathcal{F}$. Given the camera $\mathcal{C}^i \in \mathcal{C}$ and a projection function (\textit{e.g.} a renderer~\cite{Kato_2018_CVPR}) the mesh in rendered on the camera $\mathcal{C}^i$ producing a data structure $F^i \in \mathbb{R}^{w \times h}$ (\textit{i.e.} projecting the faces onto the image plane of the camera $\mathcal{C}^i$) and the rendering of the 3D body mesh 
$\mathcal{M}^i$ (Fig~\ref{fig:mods_}(b)).

\noindent\textbf{Human part segmentation.} 
Let us decompose the human body representation into $B$ parts (\textit{e.g.} head and arms).
Since for the SMPL the map $\mathcal{F}$ has a fix set of faces, we cluster it into parts\footnote{In practice, we manually annotate each of the $N_f$ face into a unique body-part label.} such that: $\mathcal{F} = \{ \mathcal{F}_b \}_{b=1}^B$. Therefore, each face $f \in F^i$ can belong to any of the $B$ classes or to the background.

\noindent\textbf{Foreground motion.}
We exploit the data-structure $\{F^i_t\}_{t = 1}^T$ to extract the foreground motion information.
Specifically, because mesh vertices are uniquely identified over time, we can compute their displacement 3D vector. 
This 3D vector can be projected on the image to obtain the foreground motion flow.
As in~\cite{Li_2018_ECCV} we use a backward motion flow to warp the frame for each time step to help the foreground synthesis.
Given a face $f \in F^i_{t+1}$ (resp. $F^i_{t}$) at pixel location $(u_x, u_y)$ (resp. $(u_x^{'}, u_y^{'})$), let $\mathcal{O}^i_{t+1 \to t}$ be the motion vector at $(u_x, u_y)$, which is computed as $(u_x - u_x^{'}, u_y - u_y^{'})$.

\begin{figure*}[t!]
    \centering
    \begin{overpic}[width=1\textwidth]{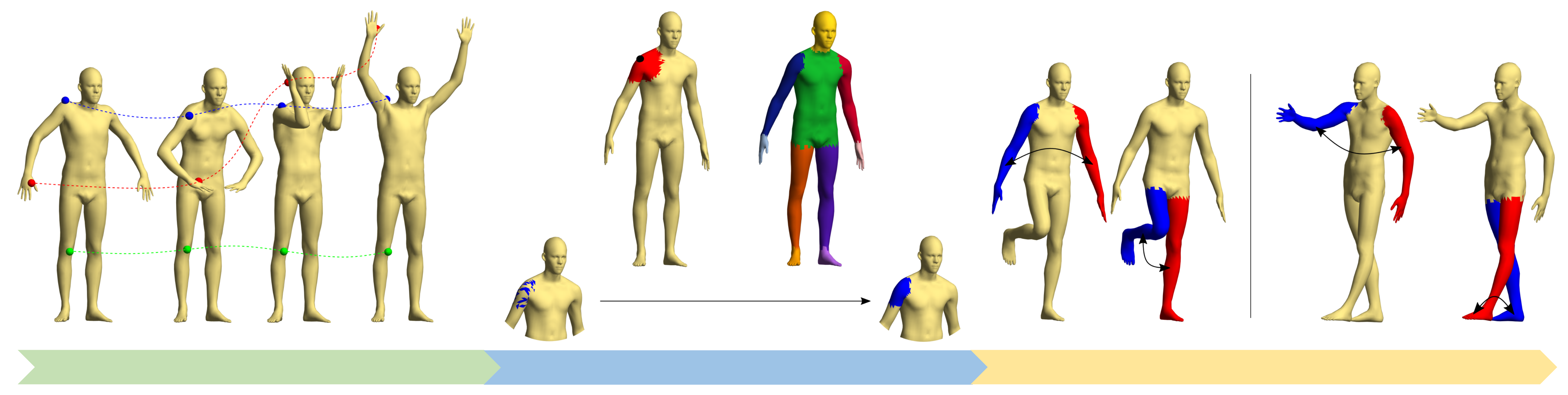}
    \put(6.4,1.4){\color{black}\tiny  Visible texture transfer}
    \put(34.4,1.4){\color{black}\tiny Transfer to neighboring faces}
    \put(65.4,1.4){\color{black}\tiny Transfer to symmetric body parts}
    \end{overpic} 
    \caption{Illustration of the proposed texture transfer. \textbf{Step I:} for every visible face index from the input view mesh, we accumulate its RGB pixel value over time. Then, we copy the pixel values to the target view mesh. \textbf{Step II:} we transfer the closest visible face with respect to a distance measure. \textbf{Step III:} we transfer texture across intrinsic symmetries, \textit{i.e.} from the blue regions to the red regions. Intrinsic symmetries are independent on the pose of the subject.}
    \label{fig:TT_}
\end{figure*} 

\noindent\textbf{Texture transfer.} 
The structure $F^i_t \in \mathbb{R}^{w \times h}$ is the projection of the 3D human mesh of the person in $x^i_t$ at time step $t$ onto the image plane of camera $\mathcal{C}^i$. A key observation to make is that we can exploit the association between $F^i_t$ and $x^i_t$ in order to estimate a rough foreground on the image plane of the camera $\mathcal{C}^j$. 
The proposed Symmetric Texture Transfer extends this idea to improve the target foreground appearance through three steps (Fig.~\ref{fig:TT_}). 
The first step consists of tracking each visible face in $F^i_t$ over time. If a face $f \in F^i_t$ is at position $(u_x, u_y)$ we copy the pixel value of $x^i_t$. The face-pixel association is then stored in a hashmap where the keys are the face number and the values are the pixels. If at time $t+k$ the face $f$ is detected we add it and at time step $T$ we keep the median of the detected pixels. 
The second step transfers pixels from the hashmap to an image indexed by $F^j$. Specifically, given a face $f \in \mathcal{F}$, we rank the neighboring faces as a function of the distance $\mathrm{\textbf{dist}}: \mathcal{F} \times \mathcal{F} \to  \mathbb{R}$ defined on the surface manifold of a template mesh.
Because the Euclidean distance is not a suitable metric to measure distances of vertices on a deformable surface, we use the geodesic distance \cite{Surazhsky_2005_TOG} that is invariant to intrinsic deformation of the mesh. 
The computation of the geodesic distances produces the matrix $\mathbf{F} \in \mathbb{R}^{M \times M}$, where the element in the $u^{th}$ row and and $v^{th}$ column is $\mathbf{F}_{uv} = \{ \mathrm{\textbf{dist}}(f, f') | f, f' \in \mathcal{F}\}$.
Using $\mathbf{F}$ we transfer the texture of the $n$-nearest neighbor face to the image. 
The final step uses symmetry between body part in order to transfer occluded pixels (see Supplementary Material).

We use a template gender neutral 3D mesh with a canonical pose to compute the pairwise distance map $\mathbf{F} \in \mathbb{R}^{N_f \times N_f}$ (see Fig.~\ref{fig:TT_}(a)). The reason is that for Euclidean distance it is computationally not possible to compute it for each frame of the dataset. Furthermore, computing a geodesic distance is much more computationally expensive than an Euclidean distance. The texture transfer $\mathcal{T}^s_{i \to j}$ approximates the foreground of the novel-view and is not used as a final prediction. 

\subsection{Geometric texture Transfer Network (GTNet)}
\label{sec:model}

\begin{figure}[t]
\begin{center}
\includegraphics[width=1.\columnwidth]{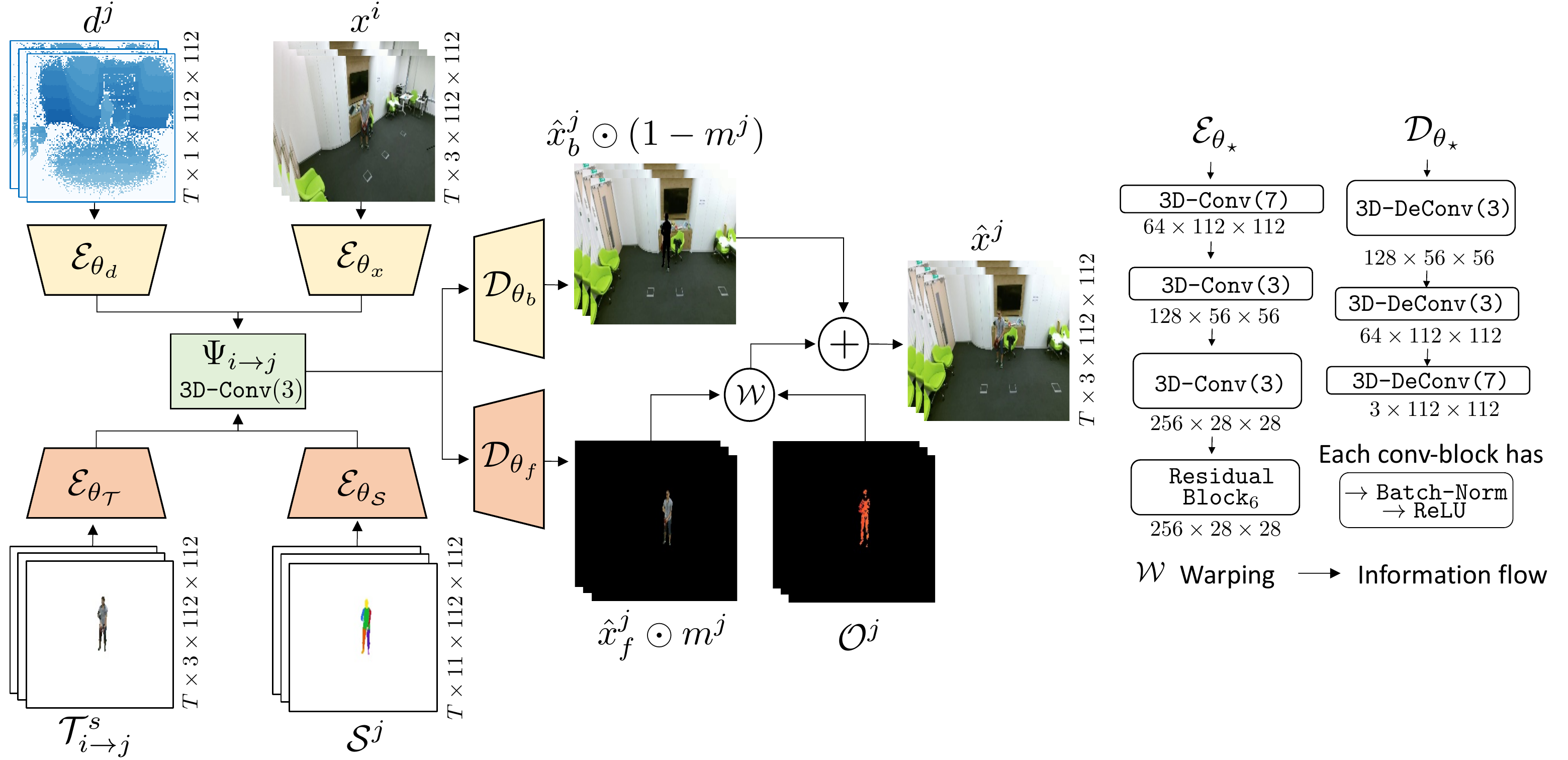}
\end{center} 
\caption{Architecture of the proposed GTNet model. We encode each modality using a separate encoder to approximate the feature point of the target view with $\Psi_{i \to j}$. We decode the background and the foreground separately. Note that we also enforce explicit temporal modeling using the estimated foreground motion.}
\label{fig:gtnet_pp}
\end{figure} 

\begin{figure}
    \centering
    \begin{overpic}[width=1.\textwidth]{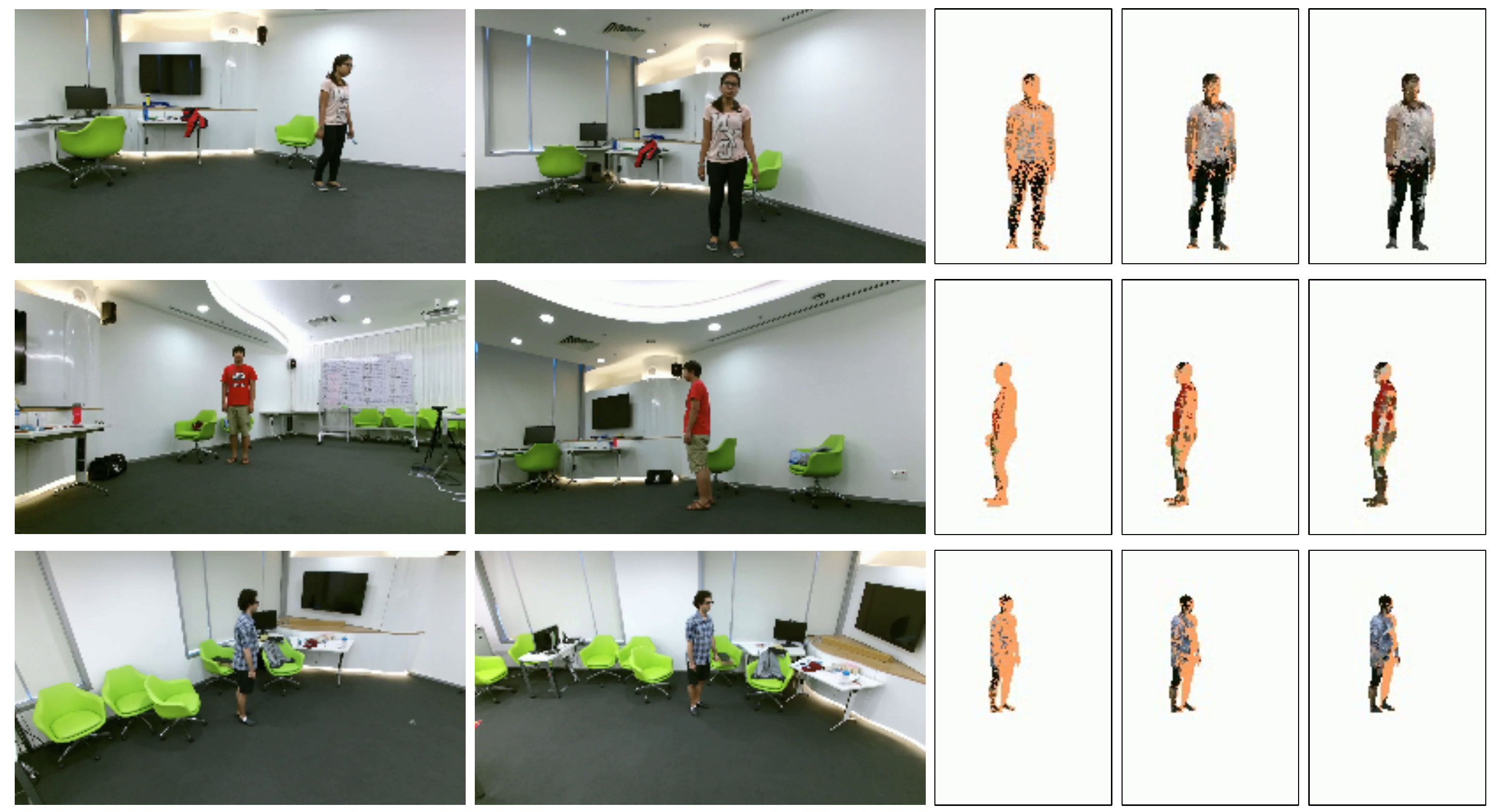}
    \put(15,54.5){\color{black}\scriptsize $x^i$}
    \put(46,54.5){\color{black}\scriptsize $x^j$}
    \put(65.5,54.5){\color{black}\scriptsize $n = 1$}
    \put(77,54.5){\color{black}\scriptsize $n = 10$}
    \put(89.5,54.5){\color{black}\scriptsize $n = 50$}
    \end{overpic} 
    \caption{Comparison of the traditional texture transfer methods (e.g.~\cite{Liu_2019_ICCV}) and the proposed transfer method  with $n \in \{10, 50 \}$ nearest neighbor transfer.}
    \label{fig:TT_cmp}
\end{figure}

The burden over the network to synthesize the novel view can be mitigated if we exploit the 3D mesh. 
The texture-transfer $\mathcal{T}^s_{i \to j}$ provides a good estimate of the foreground of target view $x^j$. Therefore we consider that $\mathcal{T}^s_{i \to j}$  is an informative input to the network and we only need to learn the residual to correct elements like mis-transferred textures or lighting. Since the texture $\mathcal{T}^s_{i \to j}$ is a good approximation of the foreground, we chose a context-based network structure. 

\noindent \textbf{Architecture.} 
GTNet jointly learns to synthesize the foreground and background (see Fig~\ref{fig:gtnet_pp}). The network takes the input view video $x^i$ (resp. depth modality $d^j$) and encodes it with feature mapping $\mathcal{E}_{\theta_x}$ (resp. $\mathcal{E}_{\theta_d}$). These features constitute the background information. Similarly, we encode the texture transfer $\mathcal{T}^s_{i \to j}$ (resp. segmentation map $\mathcal{S}^s$) to represent the foreground information in the latent space. Now to approximate the target feature $\epsilon^j$ using the operator $\Psi_{i \to j}$ we rely on a 3D Convolutional Neural Network layer, this will enforce the temporal consistency on the bottleneck layer. We therefore have the following:
\begin{equation}
    \hat{\epsilon}^j \approx \Psi_{i \to j}(\oplus_{k \in \mathcal{I}} \mathcal{E}_{\theta_{k}}(k)); \mathcal{I} = \{x^i, d^j, \mathcal{S}^j, \mathcal{T}^s_{i \to j} \},
\end{equation}
where $\oplus$ is the concatenation operation.
As motivated earlier, we separate the synthesis of the foreground and the background using dedicated decoders $\mathcal{E}_{\theta_{f}}$ and $\mathcal{E}_{\theta_{b}}$, respectively. The synthesized foreground (resp. background) is obtained as: $\hat{x}^j_{f} = \mathcal{D}_{\theta_{f}}(\hat{\epsilon}^j)$ (resp. $\hat{x}^j_{b} = \mathcal{D}_{\theta_{b}}(\hat{\epsilon}^j)$). The synthesized video $\hat{x}^j$ is therefore:
\begin{equation}
    \hat{x}^j = \hat{x}^j_{f} \odot m^j + \hat{x}^j_{b} \odot (1 - m^j),
\end{equation}
where $\odot$ is the Hadamard product and $m^j$ is the foreground mask obtained by the binarization of $F^j$.

In order to enforce temporal constraints, we propose to use the foreground motion from the mesh displacement vectors in the synthesized frame to add motion information. The frame synthesis of view $j$ at time step $t$ is defined as:
\begin{equation} \label{eq:flow_synth}
\hat{x}_{f,t}^j = 
  \begin{cases} 
   \hat{x}_{f,t}^j & \text{if } t = 1 \\
   \tilde{x}_{f,t}^j + \zeta .\mathcal{W}(\hat{x}_{f, t-1}^j, O_{t+1 \to t}^j)   & \text{if } t \in [2..T],
  \end{cases}
\end{equation}
where $\mathcal{W}$ is a residual warping function, $\hat{x}_{f,t}^j$ (resp. $O_{t+1 \to t}^j$) is the foreground prediction (resp. foreground motion) of the view $j$. $\tilde{x}_{f,t}^j$ is the initial synthesized frame of the generator and $\zeta$ is a controlling factor defined empirically (Tab.~\ref{tab:zeta}). We force the model to focus on the residue with respect to the previous time step $t-1$. Note that when training a generator, $\mathcal{W}(\hat{x}_{f, t-1}^j, O_{t+1 \to t}^j)$ is computed by a forward pass (and freezing the weights). 
Thus when applying the reconstruction pixel-wise loss, the network would only learn the residual over $\tilde{x}_{f,t}^j$.

\noindent \textbf{Training losses.} 
Instead of the traditional $L_1$ used in the literature, we employ a Huber loss~\cite{Huber_1964_AMS} to penalize the video synthesis produced by the generator. 
Differently from the $L_2$, the Huber loss is more robust to outliers and, differently from the $L_1$ loss, the Huber loss considers the directions of the error magnitude.
The reconstruction loss $L_r$ between the generated videos (both foreground and background branch) $\hat{x}^j$ and the ground-truth $x^j$ at $t$ is

\begin{equation} \label{eq:huber}
L_r =
\begin{cases}
        0.5 (\hat{x}^j_t, - x^j_t)^2, & \text{if } |\hat{x}^j_t, - x^j_t| < 1 \\
        |\hat{x}^j_t, - x^j_t| - 0.5, & \text{otherwise }
\end{cases}
\end{equation}
\noindent To enforce the perceptual quality over the generated videos we use the temporal perceptual loss~\cite{Lakhal_2019_ICCV}.
This loss extends the so-called perceptual loss~\cite{Johnson_2016_ECCV} by penalising the generated videos on a spatio-temporal feature space using a 3D CNN network $\phi$ (called perceptual network). The temporal perceptual loss is defined as:
\begin{equation} \label{eq:temp_loss}
    {L}_{p} = \displaystyle \sum_{k=1}^L \frac{1}{T_k w_k h_k c_k} \norm{  \phi_k(\hat{x}^j) - \phi_k(x^j) }_2,
\end{equation}
where $T_k, w_k, h_k, c_k$ are the temporal dimension (\textit{i.e. timesteps}), width, height and the number of channel of at the $k$-th layer of the perceptual network $\phi$, respectively.
Furthermore, we use adversarial loss~\cite{Goodfellow_2014_NIPS} in order to add high frequency details in the synthesized frames. Given our generator $G_{i \to j}$ and a discriminator $D$, the conditional adversarial loss is given as:

\begin{equation}
  L_a = \mathbb{E}_{x^i, x^j}\Big[ \log(D(x^i, x^j)) \Big] + \mathbb{E}_{x^i}\Big[ \log(1 - D(x^i, \hat{x}^j)) \Big].
\end{equation}
The total training loss is given by $L = L_r + \lambda_p L_p + \lambda_a L_a$, with $\lambda_p = \lambda_a = 0.01$.

%% file: sec_exp.tex
\section{Experiments}
This section evaluates the proposed GTNet. Sec.~\ref{s6_setup} describes the training protocol. Sec.~\ref{s6_ablation} provides the ablation of each component of the proposed pipeline. Sec.~\ref{s6_comp} compares our method with the state-of-the art VDNet~\cite{Lakhal_2019_ICCV}.

\begin{figure}[t!]
    \centering
    \begin{overpic}[width=1.\textwidth]{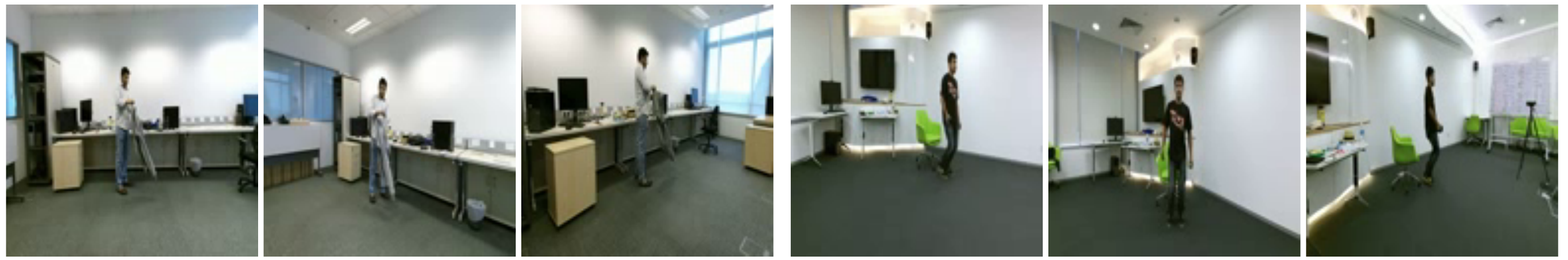} 
    \put(7,17){\color{black}\scriptsize $\mathcal{C}^1$}
    \put(24,17){\color{black}\scriptsize $\mathcal{C}^3$}
    \put(41,17){\color{black}\scriptsize $\mathcal{C}^3$}
    
    \put(57.5,17){\color{black}\scriptsize $\mathcal{C}^1$}
    \put(74.5,17){\color{black}\scriptsize $\mathcal{C}^3$}
    \put(90,17){\color{black}\scriptsize $\mathcal{C}^3$}
    \end{overpic}
    \caption{Sample frames from the NTU RGB+D dataset. The 3 views are captured with cameras placed with horizontal angle of: $-45^{\circ}, 0^{\circ}, +45^{\circ}$.}
    \label{fig:ntu_samples}
\end{figure}
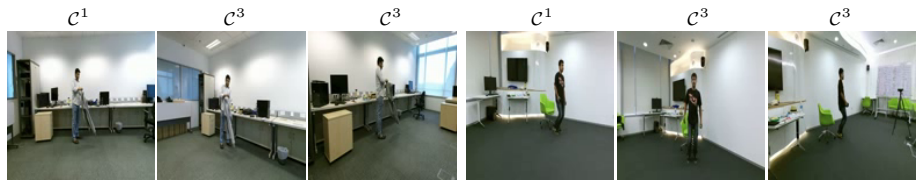

\subsection{Experimental Setup} \label{s6_setup} 
\noindent \textbf{Dataset.} We use NTU RGB+D~\cite{Shahroudy_2016_CVPR}, the only large-scale synchronized multi-view action recognition dataset (see Fig.~\ref{fig:ntu_samples}), which consists of of videos captured using three synchronized cameras with two front views and one side view. The dataset contains 80 views with 40 distinct subjects and 60 actions. Following~\cite{Lakhal_2019_ICCV}, we use the cross-subject split.\\
\noindent \textbf{Evaluation metrics.} We assess the performance using two criteria: (i) the generated video visual quality; (ii) the accuracy of the pose of the individual. For the visual quality, we use Structural Similarity (SSIM), Peak Signal-to-Noise-Ratio (PSNR)~\cite{Wang_2004_TIP} (we also report their masked version~\cite{Ma_2017_NIPS}) and Fr\'echet Video Distance (FVD)~\cite{Unterthiner_2019_ICLRW}. We use Percentage of Correct Keypoints (PCK)~\cite{Yang_2013_PAMI} for the pose evalutation.

\begin{figure}[t!]
\begin{minipage}{\textwidth}
    \begin{minipage}[b]{0.40\textwidth}
   \label{fig:model_abl}
    \centering 
    \includegraphics[width=1.\textwidth]{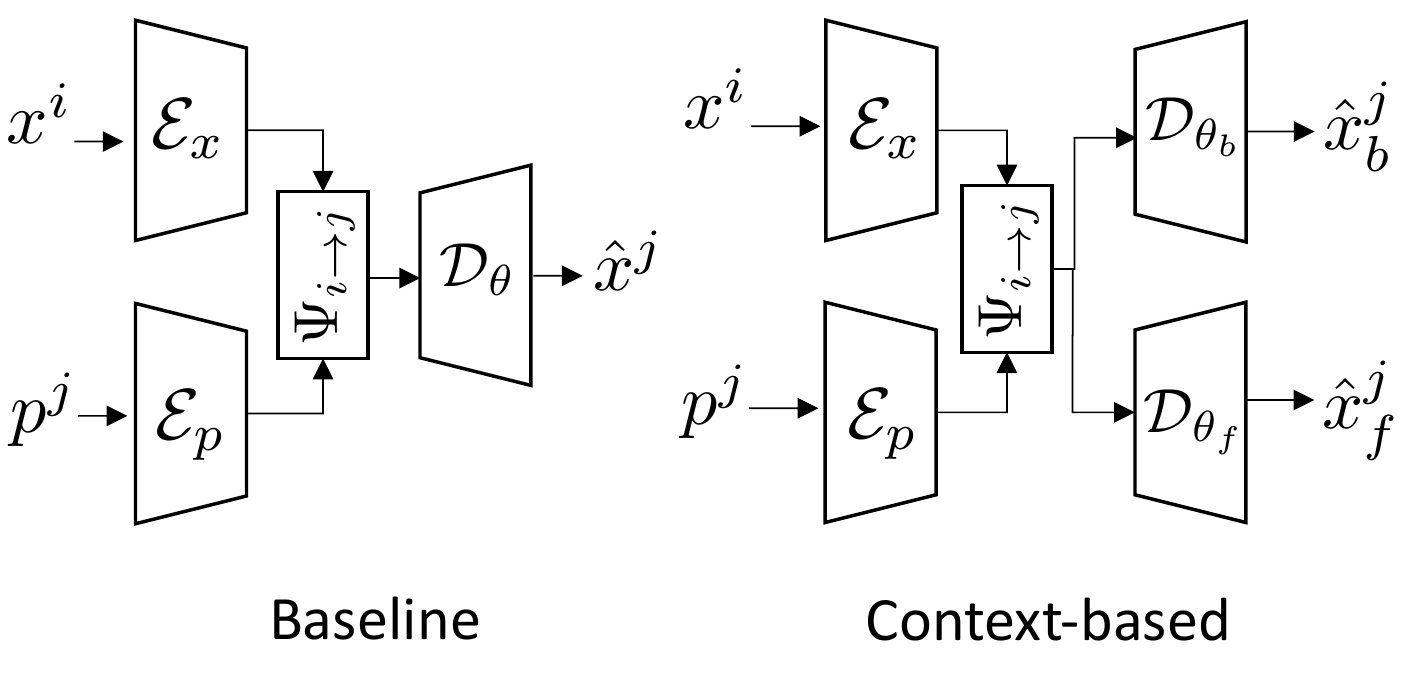} 
    \captionof{figure}{Model ablation.}
  \end{minipage}
  \begin{minipage}[b]{0.58\textwidth}
    \centering
  \includegraphics[width=1.\textwidth]{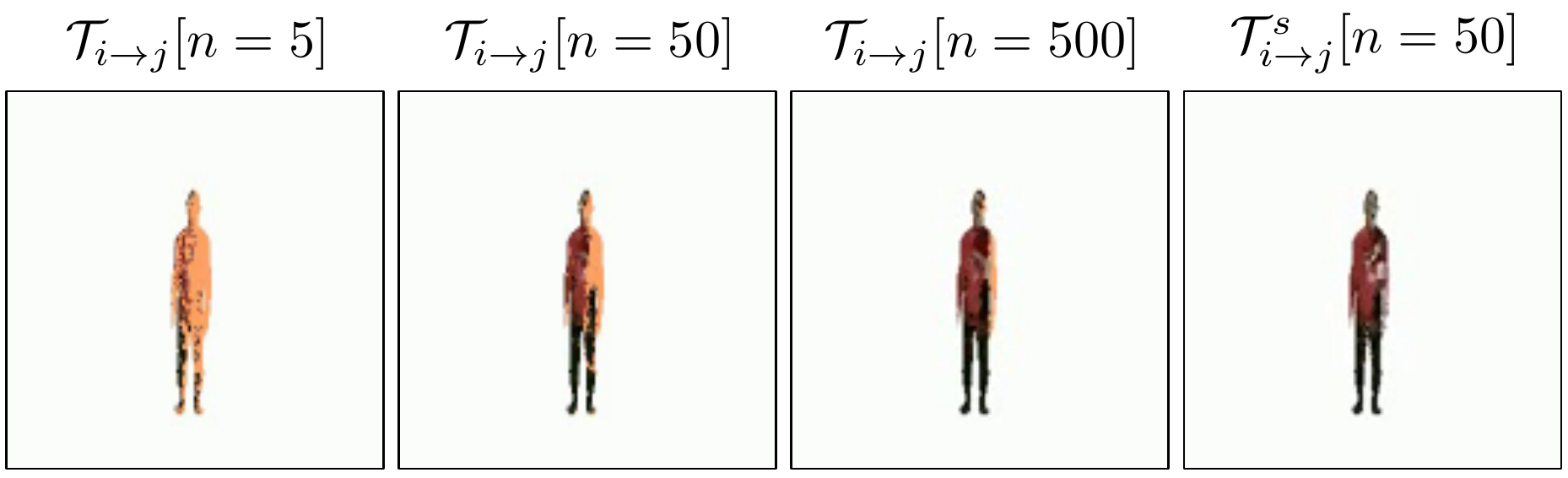} 
    \captionof{figure}{Texture transfer vs. Symmetric texture transfer (occluded region: orange pixel).}
  \end{minipage}
  \end{minipage}
 \end{figure}

\noindent \textbf{Implementation Details.}
To obtain a temporally consistent 3D mesh we combine OpenPose~\cite{Raaj_2019_CVPR} with~\cite{Kanazawa_2019_CVPR}.
We used an NVIDIA Tesla V100 16GB RAM GPU to train our model. 
We use Adam optimizer~\cite{Kingma_2015_ICLR} with $(\alpha_1, \alpha_2)=(0.5, 0.999)$ and a learning rate of 2$\cdot$10$^{-5}$.
\subsection{Ablation Studies} \label{s6_ablation} We provide a detailed  evaluation of each component of the proposed pipeline. Unless otherwise stated, we use a 2D-ResNet$_6$~\cite{Zhu_2017_ICCV} for the ablation.\\
\noindent \textbf{Texture Transfer.} We show the result of the contribution of each step in the texture transfer:
\begin{itemize}[leftmargin=*]
  \item \textbf{Step II (Euclidean):} $\mathbf{F}$ computed with Euclidean pairwise distance. \item \textbf{Step II:} $\mathbf{F}$ computed with geodesic pairwise distance.
  \item \textbf{Step II + III:} Symmetric texture transfer.
  \item \textbf{Step I + II + III:} Symmetric texture transfer with temporal context.
\end{itemize}

\noindent Results from Tab.~1 show that the symmetric texture transfer helps to better estimate the foreground with only a kernel size of $n=50$ instead of 500 without the symmetry. Adding the temporal context for the hashmap construction improves further. Fig.~7 shows a challenging case of texture transfer between views with and without the body symmetry transfer.

\begin{figure}[t!]
\begin{minipage}{\textwidth}
\scriptsize

  \begin{minipage}[b]{0.41\textwidth}
  \label{tab:tt_cmp}
      \captionof{table}{Quality of the foreground estimation. Key. M: mask; S.: SSIM; P. PSNR; Euc: Euclidean; []: nearest neighbour value $n$.}
  \scriptsize
    \centering
    \begin{tabular}{lccc}
        \hline
        \textbf{Step.} & \textbf{Notation.} & \textbf{M-S.} & \textbf{M-P.} \\ \hline
        II (Euc.) & \cellcolor{white}  $\mathcal{T}_{i \to j}[500]$ & \cellcolor{white} $.952$ & \cellcolor{white} $26.85$\\
    II & \cellcolor{Gray} $\mathcal{T}_{i \to j}[500]$ & \cellcolor{Gray} $.952$ & \cellcolor{Gray} $26.86$ \\
    II, III & \cellcolor{white} $\mathcal{T}^s_{i \to j}[50]$ & \cellcolor{white} $.953$ & \cellcolor{white} $26.92$  \\
    I, II, III & \cellcolor{Gray} $\mathcal{T}^s_{i \to j}[50]$ & \cellcolor{Gray} $.954$ & \cellcolor{Gray} $27.16$\\
        \hline
    \end{tabular}
    \end{minipage}
  \begin{minipage}[b]{0.51\textwidth}
  \label{fig:abl_base}
  \captionof{table}{Baseline ablation. Key. M: mask; S.: SSIM; P. PSNR; Mod. modality; BL: baseline; Hb: Huber.}
    \centering
    \begin{tabular}{l|cccccc} \hline
    \rowcolor{white}
    \textbf{Model} & \textbf{Mod.} & \textbf{S.} & \textbf{M-S.} & \textbf{P.} & \textbf{M-P.} & \textbf{FVD} \\
    \hline
    BL ($\Psi^{\texttt{lin}}_{i \to j}$) &  \cellcolor{white} $\mathcal{M}^j_{2D}$ &    \cellcolor{white} $.534$ &  \cellcolor{white} $.957$ &  \cellcolor{white} $17.62$ &  \cellcolor{white} $26.13$ &  \cellcolor{white} $10.81$ \\ \hline
    \multirow{3}{*}{BL ($\Psi^{\texttt{conv}}_{i \to j}$)} & \cellcolor{Gray} $\mathcal{M}^j_{2D}$  & \cellcolor{Gray} $.628$ & \cellcolor{Gray} $.964$ & \cellcolor{Gray} $18.39$ & \cellcolor{Gray} $27.73$ & \cellcolor{Gray} $7.51$ \\
     & \cellcolor{white}  $\mathcal{T}_{i \to j}$ &  \cellcolor{white} $.680$ & \cellcolor{white} $.969$ & \cellcolor{white} $19.83$ & \cellcolor{white} $29.13$ & \cellcolor{white} 6.79 \\
     & \cellcolor{Gray} $\mathcal{T}^s_{i \to j}$ &  \cellcolor{Gray} $.688$ & \cellcolor{Gray} $.970$ & \cellcolor{Gray} $19.85$ & \cellcolor{Gray} $29.12$ & \cellcolor{Gray} 6.57 \\
    \hline
    GTNet ($L_1$) & \cellcolor{white} $\mathcal{T}_{i \to j}$  & \cellcolor{white} $.693$ & \cellcolor{white} $.977$ & \cellcolor{white} $20.26$ & \cellcolor{white} $31.81$ & \cellcolor{white} 6.81 \\ 
   GTNet (Hb) & \cellcolor{Gray} $\mathcal{T}_{i \to j}$&  \cellcolor{Gray} $.709$ & \cellcolor{Gray} $.976$ & \cellcolor{Gray} $20.63$ & \cellcolor{Gray} 31.70 & \cellcolor{Gray} $6.44$ \\ \hline
\end{tabular}
    \end{minipage}
  
  \end{minipage}
 \end{figure}

\begin{figure}[t!]

\begin{minipage}{\textwidth}
\begin{minipage}[b]{0.43\textwidth}
  \scriptsize
  \captionof{table}{Sensitivity analysis of \\the warping factor with $T = 24$.}
    \label{tab:zeta}
    \centering
    \begin{tabular}{ccccccc}
        \hline
        $\zeta$ & $0$& $.1$ & $.01$ & $.001$\\ \hline
        \rowcolor{Gray}
        \textbf{SSIM}  & $.624$ & $\textbf{.635}$ & $.623$ & $.612$ \\
        \textbf{PSNR}  & $18.35$ & $\textbf{18.41}$ & $18.17$ & $18.19$ \\
        \hline
    \end{tabular}
    \end{minipage}
  \begin{minipage}[b]{0.55\textwidth}
    \centering
\scriptsize

      \captionof{table}{Synthesis performance using different model weight using $T = 8$.}
\centering
\label{tab:params}
\begin{tabular}{lccc|cc}
\hline
& \multicolumn{3}{c}{VDNet~\cite{Lakhal_2019_ICCV}} & \multicolumn{2}{c}{GTNet} \\ \hline
\rowcolor{Gray}
$\#$layers & $6$ ($3$D) & $6$ & $18$ & $6$ & $6$ ($3$D)\\
$\#$params & $112.74$M & $34.70$M & $77.20$M & $12.35$M & $99.20$M \\
\rowcolor{Gray}
\textbf{SSIM} & $.821$ & $.698$ & $.711$ & $.709$ & $\textbf{.823}$ \\
\textbf{M-SSIM} & $.972$ & N/A & N/A & $.976$ & $\textbf{.981}$ \\
\hline
\end{tabular}
\end{minipage}
  \end{minipage}
\end{figure}

\begin{figure}
    \centering
    \includegraphics[width=1.\textwidth]{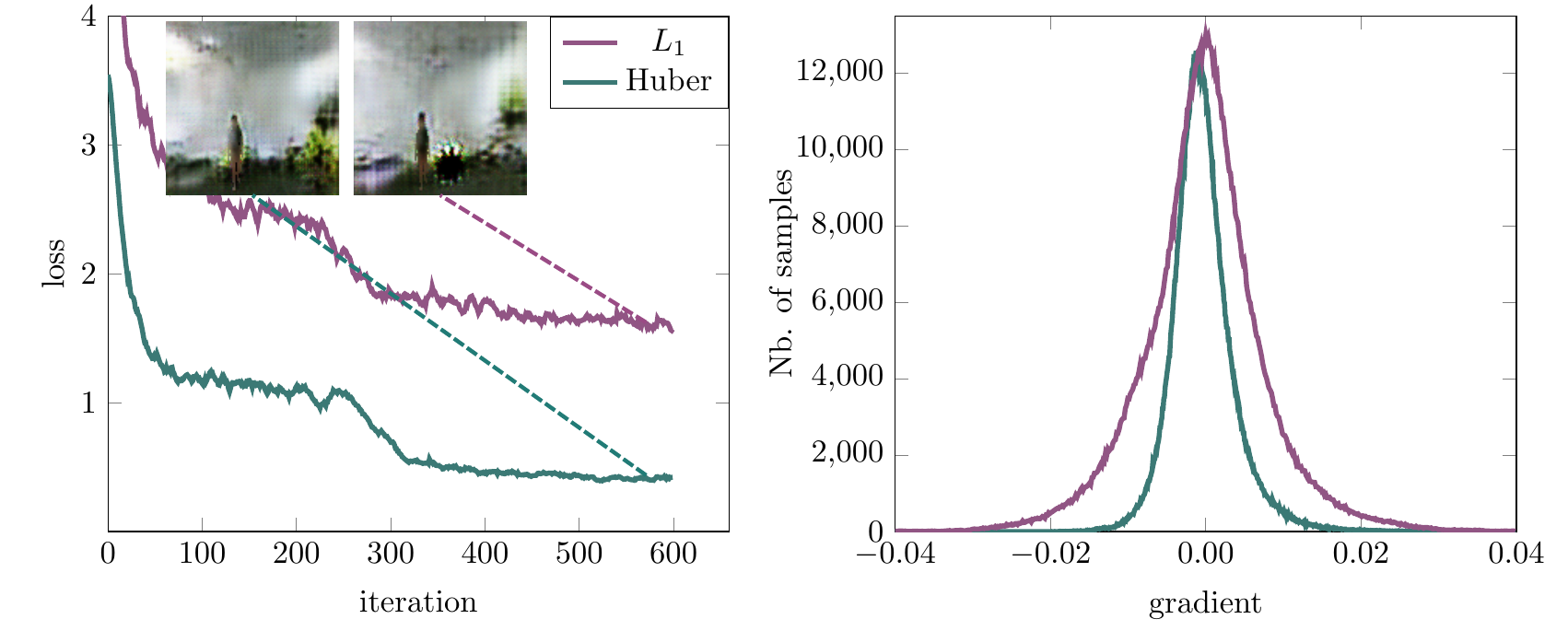} \caption{Training curves analysis of GTNet.}
    \label{fig:loss_}
\end{figure}

\begin{table}[b!]
		\captionof{table}{Models performance using a pose estimator~\cite{Raaj_2019_CVPR}.}
    \centering
    \scriptsize
    \label{tab:pck}
    \begin{tabular}{lcccc}
        \hline
        \multirow{2}{*}{\textbf{Model}} & \multirow{2}{*}{$\textbf{L}_2$} & \multicolumn{3}{c}{\textbf{PCK} \cite{Yang_2013_PAMI}} \\ 
        && 0.20 & 0.05 & 0.01\\ \hline
        \rowcolor{Gray}
        VDNet~\cite{Lakhal_2019_ICCV}  & $4.37$ & $99.3$ & $92.4$ & $51.2$ \\
        \rowcolor{white}
        Baseline  & $4.06$ & $99.4$ & $\textbf{93.6}$ & $55.3$ \\
        \rowcolor{Gray}
        GTNet  & $\textbf{3.95}$ & $\textbf{99.5}$ & $93.0$ & $\textbf{57.6}$ \\
        \hline
    \end{tabular}
\end{table}

\noindent \textbf{Baseline Models.} GTNet has a separate decoder for the foreground and the background. Therefore, we chose a generator with a single decoder as the baseline (see Fig.~6). GTNet is key to refine the texture transfer. To verify this we consider three variants of the input to the network: $\mathcal{M}^j_{2D}, \mathcal{T}_{i \to j}$, and   $\mathcal{T}^s_{i \to j}$.

We propose two variants of  $\Psi_{i \to j}$ to assess the feature approximation:
\begin{itemize}[leftmargin=*]
  \item \textbf{Linear:} $\Psi^{\texttt{lin}}_{i \to j}(\epsilon^i, \pi^j) = \mathbf{W}_{ij} . \epsilon^i +  \mathbf{W}_{jj} . \pi^j + b_j$ s.t $\mathbf{W}_{ij}, \mathbf{W}_{jj} \in \mathbb{R}^{m \times m}, b_j \in \mathbb{R}^{m}$.
  \item \textbf{Convolution:} $\Psi^{\texttt{conv}}_{i \to j}(\epsilon^i, \pi^j) = \texttt{conv}_{3 \times 3}(\epsilon^i \oplus \pi^j)$.
\end{itemize}

\noindent The operator $\Psi_{i \to j}$ estimates the feature vector of the target view. The linear version assumes a linearity between the input-view feature and the target-view modalities, whereas, the convolution applies a concatenation operation followed by a convolution operation which refers to a complex mapping (\textit{i.e.} non-linear) between the inputs. Results from Tab.~2 suggest that better feature approximation leads to better view synthesis. A linear mapping cannot approximate well the target feature $\epsilon^j$. The convolution is the default feature approximation for GTNet.

$\mathcal{M}^j_{2D}$ is the straightforward modality to use for the synthesis using 3D mesh. Tab.~2 shows that Baseline($\mathcal{M}^j_{2D}$) underperforms compared to Baseline($\mathcal{T}_{i \to j}$). This suggests that the texture transfer helps the network to refine the foreground. Having a better estimate (\textit{i.e.} $\mathcal{T}^s_{i \to j}$) improves further. With the Baseline the network has to focus on both synthesizing the foreground and background. Using the context based approach in GTNet helps the model to focus on the background synthesis and to refine only the foreground. The other conclusion is the texture $\mathcal{T}^s_{i \to j}$ approximates better the foreground.

\begin{figure*}[t!]
    \centering
    \begin{overpic}[width=1.\textwidth]{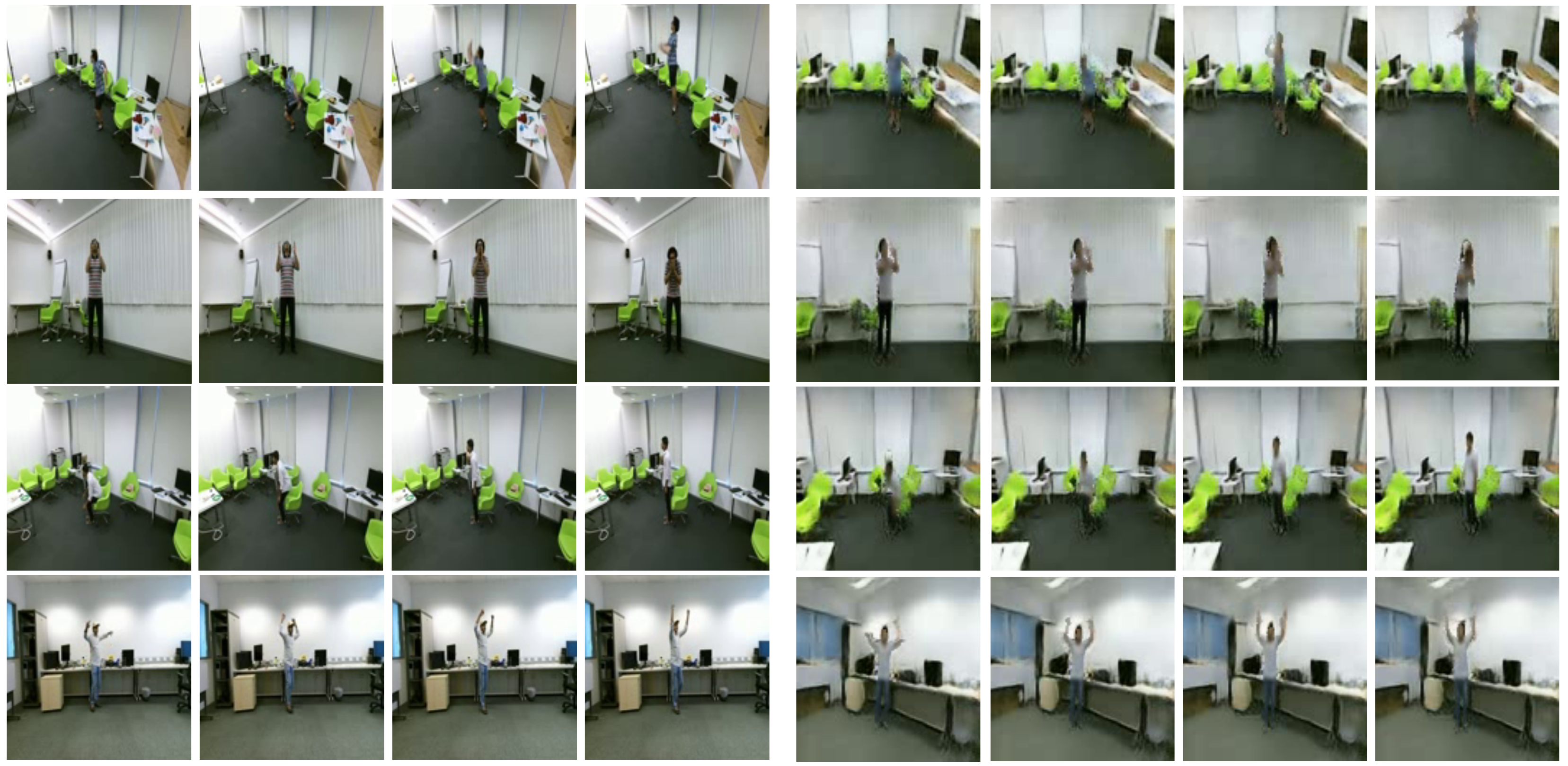}
        \put(4,49){\color{black}\scriptsize $t = 1$}
        \put(16,49){\color{black}\scriptsize $t = 8$}
        \put(28,49){\color{black}\scriptsize $t = 16$}
        \put(40,49){\color{black}\scriptsize $t = 24$}
        \put(54,49){\color{black}\scriptsize $t = 1$}
        \put(66,49){\color{black}\scriptsize $t = 8$}
        \put(78,49){\color{black}\scriptsize $t = 16$}
        \put(90,49){\color{black}\scriptsize $t = 24$}
    \end{overpic} 
    \caption{Sample frames on novel-view synthesis. (left): input view video sequence $x^i$; (right): synthesized target view $\hat{x}^j$ using the proposed GTNet.} 
    \label{fig:mv-act}
\end{figure*}

\begin{figure}[b!] 
    \centering
    \begin{overpic}[width=1.\linewidth]{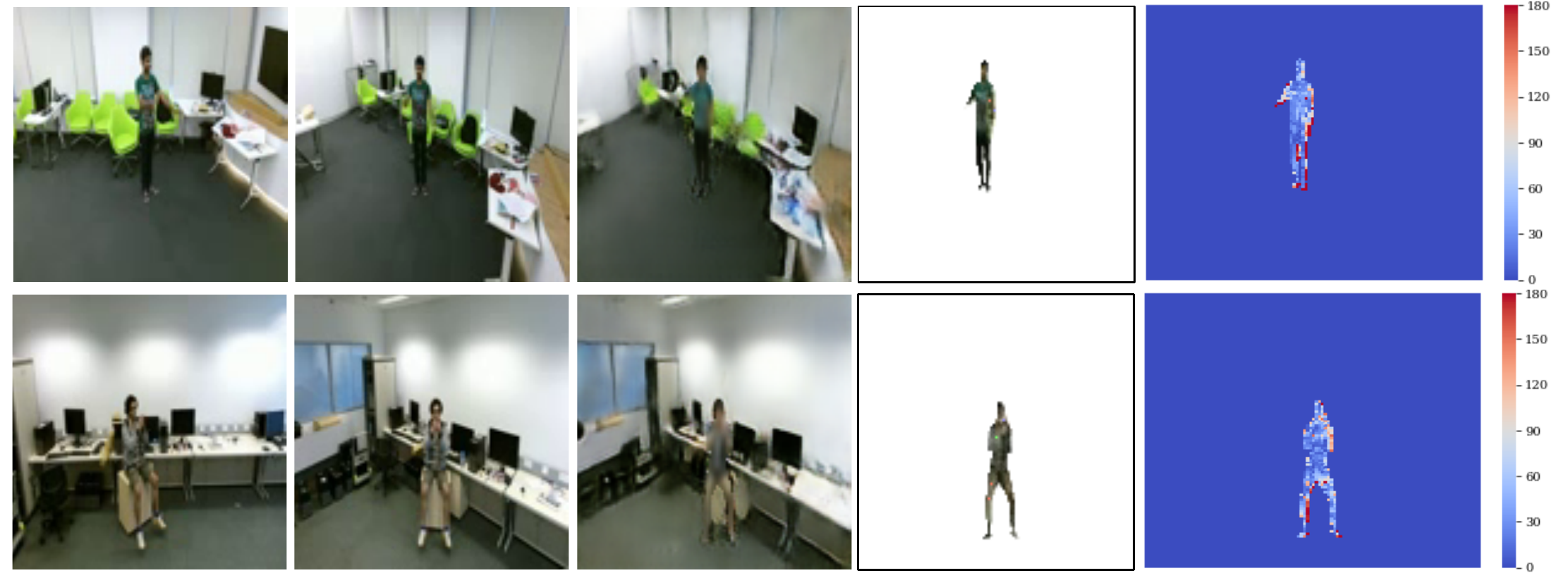}
    \put(9,37.5){\color{black}\scriptsize $x^i$}
    \put(27.5,37.5){\color{black}\scriptsize $x^j$}
    \put(45,37.5){\color{black}\scriptsize $\hat{x}^j$}
    \put(62,37.5){\color{black}\scriptsize $\mathcal{T}^s_{i \to j}$}
    \put(76,37.5){\color{black}\scriptsize $\norm{\hat{x}^j \odot m^j - x^i}$}
    \end{overpic} 
    \caption{Visualisation of the learned residual of $\mathcal{T}^s_{i \to j}$ using GTNet.}
    \label{fig:heat}
\end{figure}

\noindent \textbf{Hyperparameters.} We analyse the model performance while validating the effect of the loss, warping factor and model weights.

Using a Huber reconstruction loss in GTNet(Huber) improves the quality of the synthesized videos (see Tab.~2). We investigate this further by plotting the training loss and the histogram of the gradient at the last convolutional kernel of the decoder of GTNet (see Fig.~\ref{fig:loss_}). For the gradient we noticed a significantly smaller variance, which we deem to be due to the non-smoothness of the $L_1$ loss around the origin. By analysing the output of the generator trained with these two losses, we observe that the generator trained with $L_1$ loss outputs black artifacts during the first epochs which may cause mode collapse~\cite{Che_2016_ICLR}.

From Tab.~3 we note the improvement using the warping introduced in Eq.~\ref{eq:flow_synth}. This in fact helps the generator to only learn the residual from previous frame $\hat{x}^j_{t-1}$. We therefore keep $\zeta=.1$ as the default value for the warping function.

We report a good foreground synthesis even with a 2D-ResNet compared to VDNet. However, the network could not synthesize well the background. This is because without the temporal context the network will synthesize the background independently for each time step. With the 3D-ResNet we obtain better video synthesis compared to VDNet with many fewer trainable weights.

\begin{table}[t]
\scriptsize
\caption{Comparison between the proposed GTNet and VDNet~\cite{Lakhal_2019_ICCV} on NTU RGB+D. Key. $\zeta$ controlling factor  (see Sec.~\ref{sec:model}).}
\label{tab:cmp}
\begin{center}

\begin{tabular}{l|cccccc} \hline
    \rowcolor{white}
    \textbf{Model} & \textbf{Modality} & $\uparrow$ \textbf{SSIM} & $\uparrow$ \textbf{M-SSIM} & $\uparrow$ \textbf{PSNR} & $\uparrow$ \textbf{M-PSNR} & $\downarrow$ \textbf{FVD} \\
    \hline
    \multirow{3}{*}{VDNet~\cite{Lakhal_2019_ICCV}} & \cellcolor{Gray} $s^j$ & \cellcolor{Gray} $.749$ & \cellcolor{Gray} $.964$ & \cellcolor{Gray} $20.78$ & \cellcolor{Gray} $28.27$ & \cellcolor{Gray} $7.35$ \\
    &  \cellcolor{white} $d^j$ &  \cellcolor{white} $.794$ &  \cellcolor{white} $.970$ &  \cellcolor{white} $22.47$ &  \cellcolor{white} $29.46$ &  \cellcolor{white} $6.60$ \\
    & \cellcolor{Gray} $d^j + s^j$ & \cellcolor{Gray} $.821$ & \cellcolor{Gray} $.972$ & \cellcolor{Gray} $23.18$ & \cellcolor{Gray} $29.70$ & \cellcolor{Gray} $5.78$ \\ \hline
    GTNet($\zeta = 0$) & $\mathcal{M}^j_{2D}$ & \cellcolor{white} $.703$ & \cellcolor{white} $.976$ & \cellcolor{white} 20.16 & \cellcolor{white} 30.95 & \cellcolor{white} 6.34 \\ 
    GTNet($\zeta = 0$) & \cellcolor{Gray} $\mathcal{T}_{i \to j}$ & \cellcolor{Gray} $.767$ & \cellcolor{Gray} $.979$ & \cellcolor{Gray} 22.03 & \cellcolor{Gray} 31.98 & \cellcolor{Gray} 5.62 \\
    GTNet($\zeta = 0$) & \cellcolor{white} $\mathcal{T}_{i \to j} + \mathcal{S}^j$ & \cellcolor{white} $.714$ & \cellcolor{white} $.978$ & \cellcolor{white} $20.44$ & \cellcolor{white} $31.90$ & \cellcolor{white} 6.42 \\ 
    GTNet($\zeta = 0$) & \cellcolor{Gray} $\mathcal{T}_{i \to j} + d^j$ & \cellcolor{Gray} $.778$ & \cellcolor{Gray} $.980$ & \cellcolor{Gray} 22.96 & \cellcolor{Gray} 32.04 & \cellcolor{Gray} \textbf{4.32} \\
    GTNet($\zeta = .1$) & \cellcolor{white} $\mathcal{T}_{i \to j} + \mathcal{S}^j + d^j$ & \cellcolor{white} $.787$ & \cellcolor{white} $.980$ & \cellcolor{white} $22.98$ & \cellcolor{white} $32.25$ & \cellcolor{white} 5.06 \\
    GTNet($\zeta = .1$) & \cellcolor{Gray} $\mathcal{T}^s_{i \to j} + \mathcal{S}^j + d^j$ & \cellcolor{Gray} $\textbf{.823}$ & \cellcolor{Gray} $\textbf{.981}$ & \cellcolor{Gray} $\textbf{23.81}$ & \cellcolor{Gray} $\textbf{32.50}$ & \cellcolor{Gray} 4.96 \\ \hline
\end{tabular}
\end{center}
\end{table}

\textbf{Comparison.} \label{s6_comp} Results from Tab.~\ref{tab:cmp} are reported with 3D-ResNet$_6$. Overall, GTNet significantly improves over all the metrics compared to VDNet~\cite{Lakhal_2019_ICCV}. GTNet benefits from  the depth $d^j$ along with $\mathcal{T}_{i \to j}$. The model GTNet($\mathcal{T}^s_{i \to j} + \mathcal{S}^j + d^j; \zeta=.1$) produces superior quality results compared to VDNet.

$\mathcal{T}_{i \to j}$ is derived from the skeleton $s^j$. It is worth noting that GTNet($\mathcal{T}_{i \to j}$) is superior to VDNet($s^j$) (see Tab.~\ref{tab:cmp}). The proposed GTNet produces temporally consistent videos (FVD scores). Tab.~\ref{tab:pck} reports the PCK scores of GTNet with the baseline and VDNet. The pose estimator estimates keypoints that are close to the ground-truth with GTNet.

\begin{figure}[t!] 
    \centering
    \includegraphics[width=1.\textwidth]{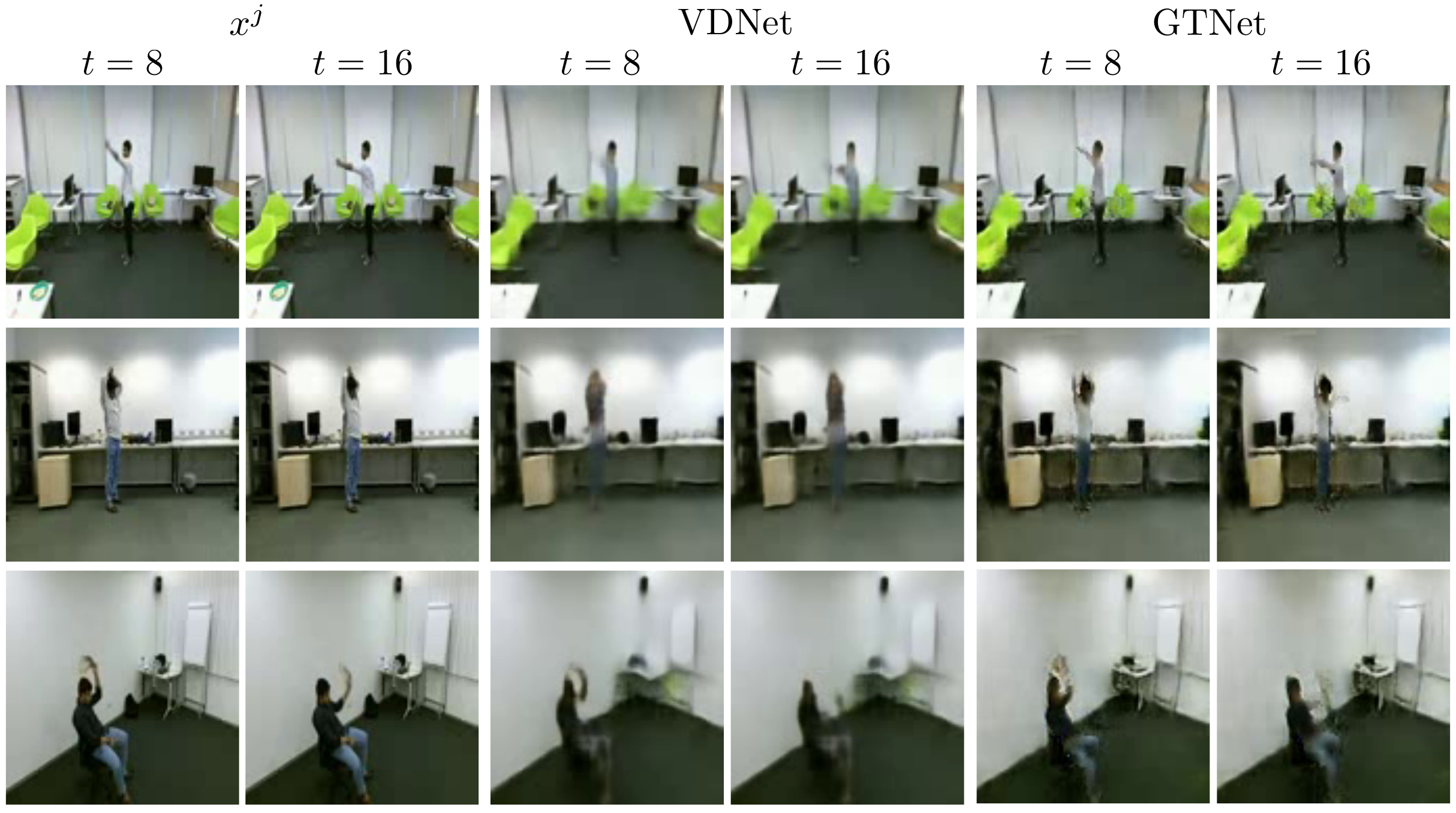} 
    \caption{Sample frames comparing VDNet and GTNet.}
    \label{fig:soa_comp} 
\end{figure}

\begin{figure}[t!]
    \centering
    \includegraphics[width=1.\linewidth]{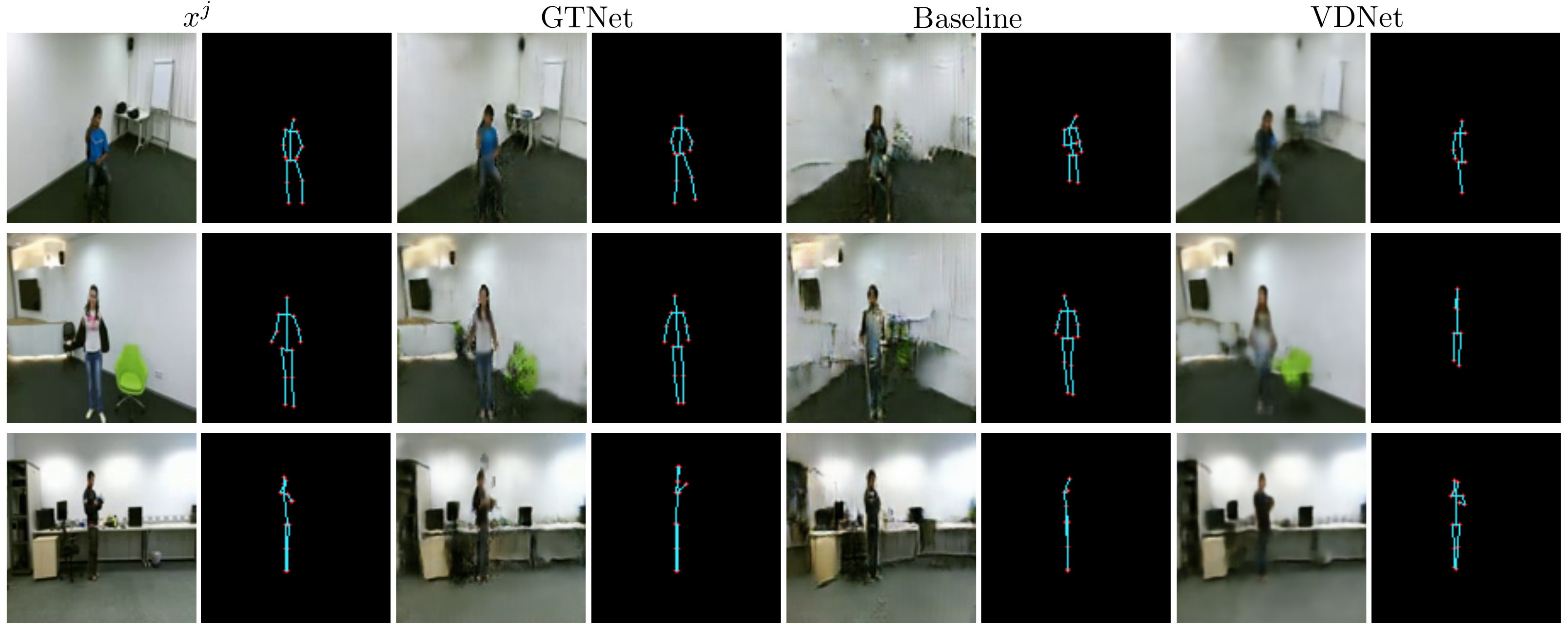}
    \caption{Comparison of the estimated skeleton.}
    \label{fig:qual_ske}
\end{figure}

Fig.~\ref{fig:heat} shows the ability of GTNet in refining the textures. Fig.~\ref{fig:mv-act} shows four examples of typical synthesis results using GTNet. The synthesized novel-view videos are sharper and we can clearly distinguish the movement of the subject. Fig.~\ref{fig:soa_comp} compares three examples of GTNet and VDNet. We can note that indeed the motion is clearly distinct with our model. Note also that thanks to $\mathcal{T}_{i \to j}^j$ the body texture is preserved across the views. In the example of the third row, we can see that GTNet is able to keep the movement of the hand with the object interaction (hat). Fig.~\ref{fig:qual_ske} shows a qualitative example of the pose estimation. The pose estimator is able to extract keypoints similar to the ones extracted from the ground-truth. This is because GTNet has better foreground synthesis ($.981$ M-SSIM, $32.50$ M-PSNR) compared to VDNet ($.972$ M-SSIM, $29.70$ M-PSNR).

%% file: sec_conclusion.tex
\section{Conclusions}
We presented a novel approach to synthesize actions as if they were recorded from a novel view by exploiting geometric and appearance information extracted from the real view.
Our geometric approach transfers the textures of the visible parts of the human (foreground) from images to a 3D mesh and re-projects them onto the novel, 2D view.
Then, we designed a new encoder-decoder network architecture that learns how to synthesize the occluded parts of the foreground and that tackles the foreground and background tasks separately to achieve high synthesis fidelity. We obtain state-of-the-art synthesis results on the NTU RGB+D dataset.

\textbf{Acknowledgements} This project acknowledges the use of the ESPRC funded Tier 2 facility, JADE.

%% file: sec_supp.tex
\section{Supplementary}
\subsection{Texture transfer}
\begin{figure}[t!]
    \centering
        \includegraphics[width=.65\textwidth]{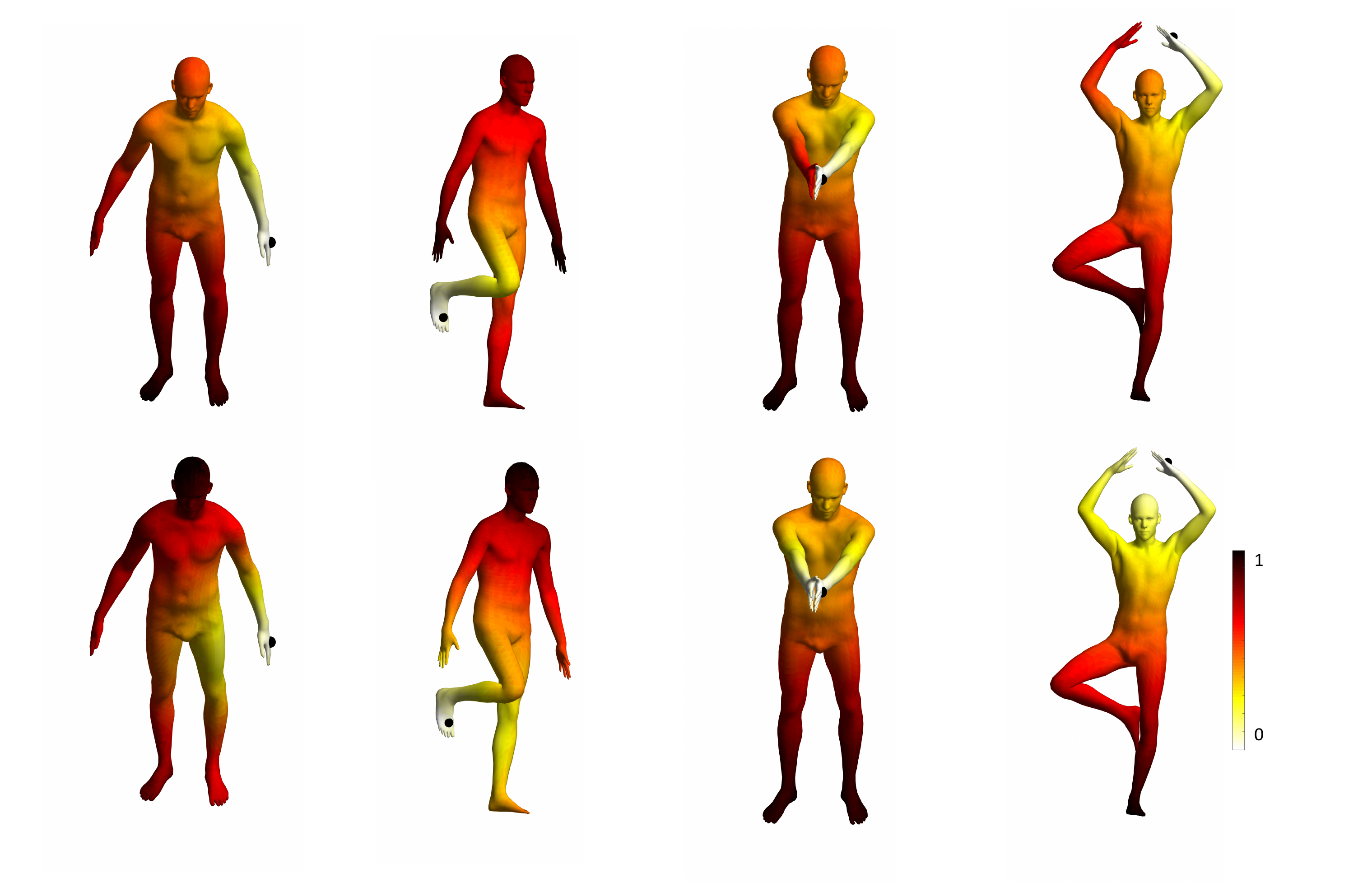}\vspace{-2em}
    \caption{Illustration of distance between a face in the mesh (black dot) with the rest of the faces. (top row): Geodesic distances; (bottom row): Euclidean distances. White represents small distance while red means large distance.} 
    \label{fig:supp_dist}
\end{figure}

\begin{figure}[t!]
    \centering
    \begin{overpic}[width=\linewidth]{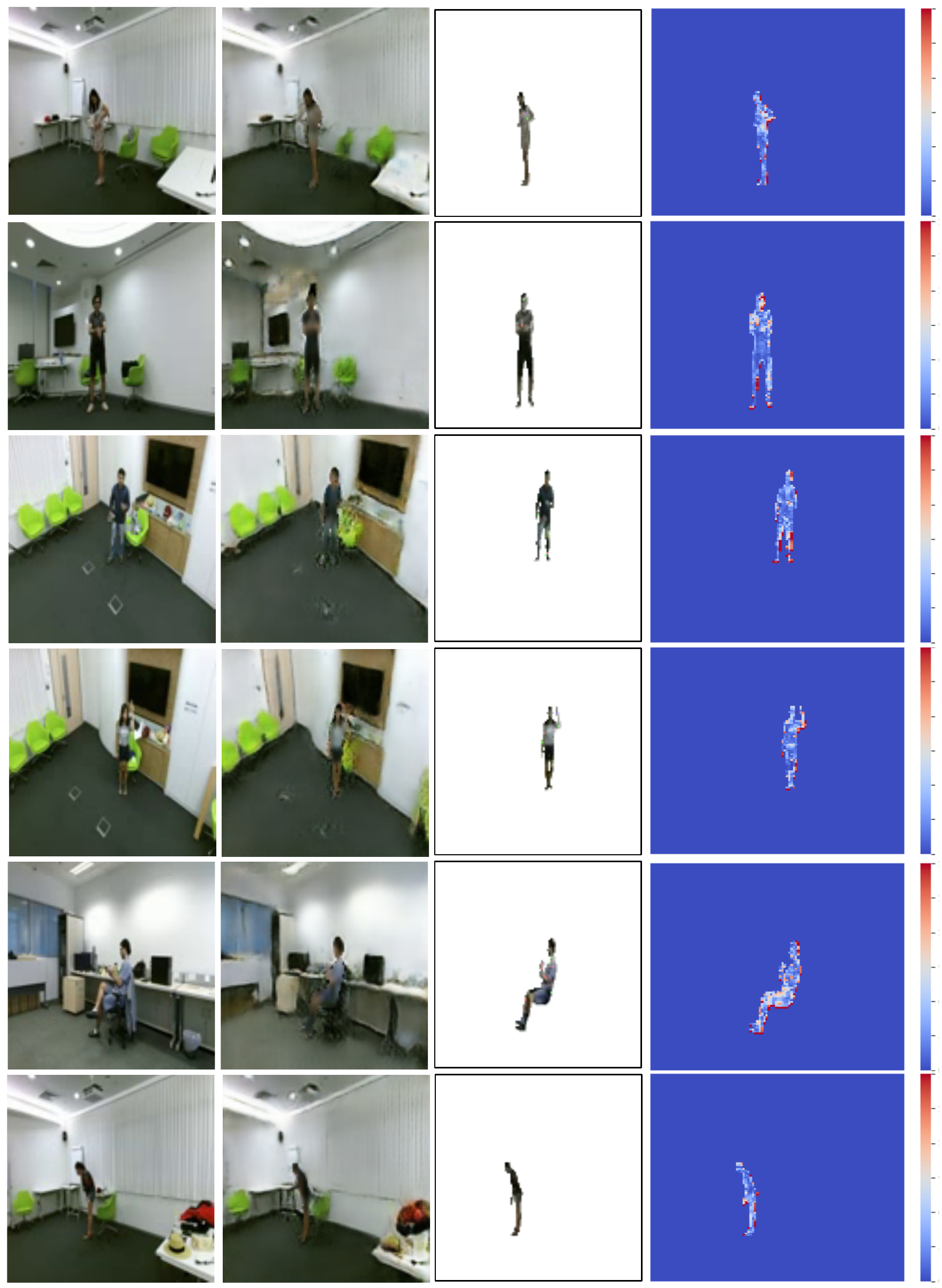}
        \put(8,100.5){\color{black}\scriptsize $x^j$}
        \put(25,100.5){\color{black}\scriptsize $\hat{x}^j$}
        \put(41,100.5){\color{black}\scriptsize $\mathcal{T}^s_{i \to j}$}
        \put(55.5,100.5){\color{black}\scriptsize $\norm{\hat{x}^j \odot m^j - x^i}$}
    \end{overpic}
    \caption{Additional results of the learned residual of $\mathcal{T}^s_{i \to j}$ using GTNet.}
    \label{fig:supp_heat}
\end{figure}

In Fig.~\ref{fig:supp_dist}, we show the map $\mathbf{F}$ computed for different body poses. We can observe that with the geodesic distance we obtain a plausible transfer (see Fig.~\ref{fig:supp_heat}). The geodesic distance encourages pixel transfer within neighboring body regions. The details of each step are presented in Alg.~\ref{alg:transfer_sym}.

\begin{algorithm}[t!]
\caption{Symmetric texture transfer}
\label{alg:transfer_sym}
  \begin{algorithmic}
    \State \textbf{Inputs:} $n < N_f$ nearest neighbor value; input-view $x^i = \{x^i_t \in \mathbb{R}^{w \times h \times 3}\}_{t = 1}^T$; time step $t$; face map $F^v_t \in \mathbb{R}^{w \times h}$ of view $v \in \{i, j\}$; $\mathbf{F} \in \mathbb{R}^{N_f \times N_f}$ pairwise face distances; \texttt{face\_to\_label} dictionary; \texttt{sym\_parts} hashmap of symmetric body parts.
    \State \textbf{Output:} Rough texture of the target view $\mathcal{T}^s_{i \to j}$\\
    \begin{minipage}[b]{0.43\linewidth}
    \scriptsize
    \Function{\texttt{step$\_$I}}{}
    \State $\mathbf{F}_{\texttt{dict}} \gets \{ \}$
      
        \For{$t \in \{1, \dots, T\}$}
            \For{$F_{\texttt{id}}, (u_x, u_y) \in F^i_t$}
                \State $\mathbf{F}_{\texttt{dict}}[F_{\texttt{id}}]\texttt{.append}(x^i_t[u_x, u_y])$
            \EndFor
        \EndFor

        \For {$F_{\texttt{id}} \in \mathbf{F}_{\texttt{dict}}\texttt{.keys()}$}
                \State $\mathbf{F}_{\texttt{dict}}[F_{\texttt{id}}] \gets \texttt{median}\{\mathbf{F}_{\texttt{dict}}[F_{\texttt{id}}]\}$
            \EndFor
    \EndFunction \vspace{1.6em}
    
    \Function{\texttt{step$\_$III}}{}
    \State $\mathcal{T}^s_{i \to j} , \texttt{d\_pix} , O_{xy} \gets \texttt{step$\_$II}()$
        
        \For{$(u_x, u_y) \in O_{xy}$}
        \State $\texttt{label} \gets \texttt{face\_to\_label}[F^i_t[u_x, u_y]]$
            \If{$\texttt{label} \not \in \texttt{d\_pix.keys()}$}
             \State $\texttt{label} \gets \texttt{sym\_parts}[\texttt{label}]$
                         \If{$\texttt{label} \not \in \texttt{d\_pix.keys()}$}
                               \State $\mathcal{T}^s_{i \to j}[u_x, u_y] \gets [\#,\#,\#]$ 
                \EndIf
                \EndIf
                
    \State $v_q \gets [u_x, u_y]$
	\State  $v_{pix}, v_{xy} \gets  \texttt{d\_pix}[\texttt{label}]$
	\State $d_L \gets L_2(v_q, v_{xy})$
    \State $\mathcal{T}^s_{i \to j}[u_x, u_y] \gets v_{pix}[\texttt{argmin}(d_L)]$  
        \EndFor
    \EndFunction
    \end{minipage}%
    \begin{minipage}[b]{0.56\linewidth}
    \scriptsize
    \Function{\texttt{step$\_$II}}{}
      \State $\mathcal{T}_{i \to j} \gets 0_{w \times h \times 3}$
     \State $\mathbf{F}_{\texttt{dict}} \gets \texttt{step$\_$I}(x^i; F^i)$
       \State $\texttt{d\_pix} \gets \{ \}$ \algorithmiccomment{label to pixel}
        \State $O_{xy} \gets []$

        \For{$F_{\texttt{id}}, (u_x, u_y) \in F^j_t$}
            \If {$F_{\texttt{id}} \in \mathbf{F}_{\texttt{dict}}\texttt{.keys()}$}
                \State $\mathcal{T}_{i \to j}[u_x, u_y] \gets \mathbf{F}_{\texttt{dict}}[F_{\texttt{id}}]$
                \State $\texttt{label} \gets \texttt{face\_to\_label}[F_{\texttt{id}}]$
                \State $F_n \gets [\mathbf{F}_{\texttt{dict}}[F_{\texttt{id}}]; (u_x, u_y)]$
                \State $\texttt{d\_pix}[\texttt{label}] \gets \texttt{d\_pix}[\texttt{label}] \cup F_n$
            \Else
                \State $\texttt{occluded} \gets \texttt{True}$
                \For{$l \in \{1, \dots, n\}$}
                    \State $F_l \gets \mathbf{F}[l_{F_{\texttt{id}}}][l]$
                    \If {$F_l \in \mathbf{F}_{\texttt{dict}}\texttt{.keys()}$}
                    \State $\mathcal{T}_{i \to j}[u_x, u_y] \gets \mathbf{F}_{\texttt{dict}}[F_l]$
                    \State $\texttt{occluded} \gets \texttt{False}$
                    \State $\texttt{label} \gets \texttt{face\_to\_label}[F_l]$
                    \State $F_n \gets [\mathbf{F}_{\texttt{dict}}[F_l]; (u_x, u_y)]$
                   \State $\texttt{d\_pix}[\texttt{label}] \gets \texttt{d\_pix}[\texttt{label}] \cup F_n$
                    \EndIf
                \EndFor
                \If {\texttt{occluded}}
                            \State $O_{xy} \gets O_{xy} \cup (u_x, u_y)$
                \EndIf
            \EndIf
            
        \EndFor
        \EndFunction
    \end{minipage}%
    \end{algorithmic}
\end{algorithm}

